\documentclass[conference]{IEEEtran}
\IEEEoverridecommandlockouts
\usepackage{cite}
\usepackage{amsmath,amssymb,amsfonts}
\usepackage{algorithmic}
\usepackage{graphicx}
\usepackage{textcomp}
\usepackage{xcolor}
\usepackage{algorithm}
\usepackage{algorithmic}
\usepackage[colorlinks,linkcolor=black]{hyperref}
\usepackage{subfigure}

\def\BibTeX{{\rm B\kern-.05em{\sc i\kern-.025em b}\kern-.08em
    T\kern-.1667em\lower.7ex\hbox{E}\kern-.125emX}}
\makeatletter
\newenvironment{breakablealgorithm}
  {% \begin{breakablealgorithm}
   \begin{center}
     \refstepcounter{algorithm}% New algorithm
     \hrule height.8pt depth0pt \kern2pt% \@fs@pre for \@fs@ruled
     \renewcommand{\caption}[2][\relax]{% Make a new \caption
       {\raggedright\textbf{\ALG@name~\thealgorithm} ##2\par}%
       \ifx\relax##1\relax % #1 is \relax
         \addcontentsline{loa}{algorithm}{\protect\numberline{\thealgorithm}##2}%
       \else % #1 is not \relax
         \addcontentsline{loa}{algorithm}{\protect\numberline{\thealgorithm}##1}%
       \fi
       \kern2pt\hrule\kern2pt
     }
  }{% \end{breakablealgorithm}
     \kern2pt\hrule\relax% \@fs@post for \@fs@ruled
   \end{center}
  }
\makeatother

\begin{document}

\title{Apply Artificial Neural Network to Solving Manpower Scheduling Problem
}

\author{\IEEEauthorblockN{1\textsuperscript{st} Tianyu Liu}
\IEEEauthorblockA{\textit{ZJU-UIUC Institute} \\
\textit{Zhejiang University}\\
Haining, China}
\IEEEauthorblockA{\textit{University of Illinois at Urbana-Champaign}\\
Urbana-Champaign, USA\\
tianyu.18@intl.zju.edu.cn}\\

\and
\IEEEauthorblockN{2\textsuperscript{nd} Lingyu Zhang}
\IEEEauthorblockA{\textit{School of Computer Science and Technology} \\
\textit{Shandong University}\\
Jinan, China}
\IEEEauthorblockA{\textit{Didi AI Labs} \\
\textit{Didi Chuxing}\\
Beijing, China\\
805906920@qq.com}\\
}

\maketitle
\begin{abstract}
The manpower scheduling problem is a kind of critical combinational optimization problem. Researching solutions to scheduling problems can improve the efficiency of companies, hospitals, and other work units. This paper proposes a new model combined with deep learning to solve the multi-shift manpower scheduling problem based on the existing research. This model first solves the objective function's optimized value according to the current constraints to find the plan of employee arrangement initially. It will then use the scheduling table generation algorithm to obtain the scheduling result in a short time. Moreover, the most prominent feature we propose is that we will use the neural network training method based on the time series to solve long-term and long-period scheduling tasks and obtain manpower arrangement. The selection criteria of the neural network and the training process are also described in this paper. We demonstrate that our model can make a precise forecast based on the improvement of neural networks. This paper also discusses the challenges in the neural network training process and obtains enlightening results after getting the arrangement plan. Our research shows that neural networks and deep learning strategies have the potential to solve similar problems effectively. 
\end{abstract}
\begin{IEEEkeywords}
Neural Network, Machine Learning, Deep Learning, Time Series Forecast, Genetic Algorithm
\end{IEEEkeywords}

\section{Introduction}
For the history of the manpower scheduling problem, in 1954, Dantzig and Fulkerson \cite{dantzig1954minimizing} gave a detailed description of the scheduling problem for the first time and proposed an algorithm to solve it. Later, different scholars from distinct fields joined the research on relative problems. Due to the different needs of different companies or units nowadays, the scheduling problem's combination of constraints and optimization conditions becomes more and more complicated. Even the generation of some constraints is inherently uncertain \cite{song2017project}. These obstacles create a challenge that must be faced in generating the results of scheduling tasks. Therefore, finding a breakthrough to optimize the solution is significant. This paper proposes a scheme that obtains training data from a scheduling model and uses deep learning methods to forecast the arrangement for manpower scheduling.
\section{Related Work}
In 2019, researchers \cite{tirk19} who study the application of traditional searching algorithm to the project scheduling problem also make some progress in solving the problem. After collecting the relevant data, the researchers proposed a problem solving model based on Pareto optimization, and ran the genetic algorithm and simulated annealing algorithm on the model, and verified a small range of efficiency improvements. However, due to the limitation of the principle, the improvement space of the traditional algorithm will be continuously reduced. The problem of scheduling optimization with artificial intelligence is becoming a more important research direction.\\
\hspace*{1em}The earliest method that can be discovered to use neural networks to deal with human resource scheduling problems comes from Hao, Lai, and Tan \cite{hao2004neural}. This paper proposes a scheme to map the traditional optimization model to the neural network and optimize it. The training process of this neural network is also unique. Compared with the scheme of measuring the error between the output result and the bool value, which uses the cost function to determine the learning result, this paper adopts a new structure, shown in Figure 1\&2 \cite{hao2004neural}, to generate the detection result that meets constraints. However, consider the efficiency, this method can only be regarded as a new research direction for the scheduling problem.\\
\begin{figure}[htbp]
\begin{minipage}{0.46\linewidth}
 \centerline{\includegraphics[width=4.0cm,height=3cm]{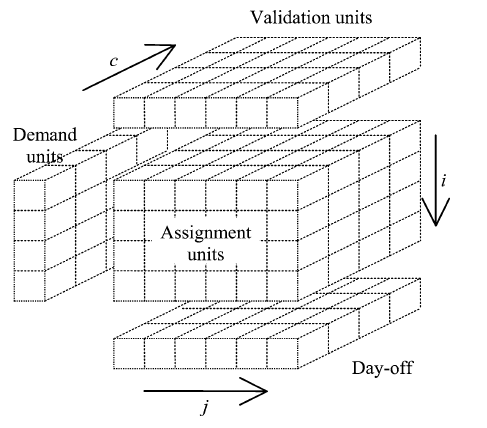}}
 \caption{Main Structure}
\end{minipage}
\begin{minipage}{0.5\linewidth}
 \centerline{\includegraphics[width=4.0cm,height=3cm]{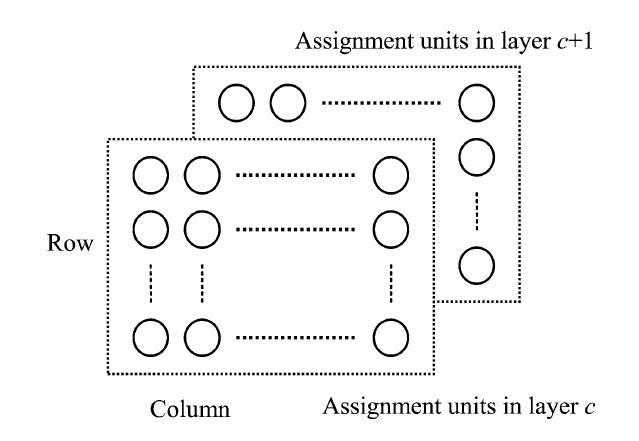}}
  \caption{Layer Structure}
\end{minipage}
\end{figure}
\hspace*{1em}A few years later, Kumari, Yadav, and Agrawal \cite{kumari19} combine genetic algorithm and convolutional neural network to create a scheme for optimizing dynamic task scheduling. Its workflow can be summarized as first applying TLBO type genetic algorithm to the data set to reduce the source set's dimension and find a preliminary feasible solution. Then researchers input the genetic algorithm's optimization results into a CNN-based neural network proposed by the authors, CNNLB, for training. Researchers prove that this choice will increase accuracy and efficiency. The flowchart proposed in this paper can be expressed as:
\begin{figure}[H] 
    	\centering
    	\includegraphics[width=5cm,height=6cm]  {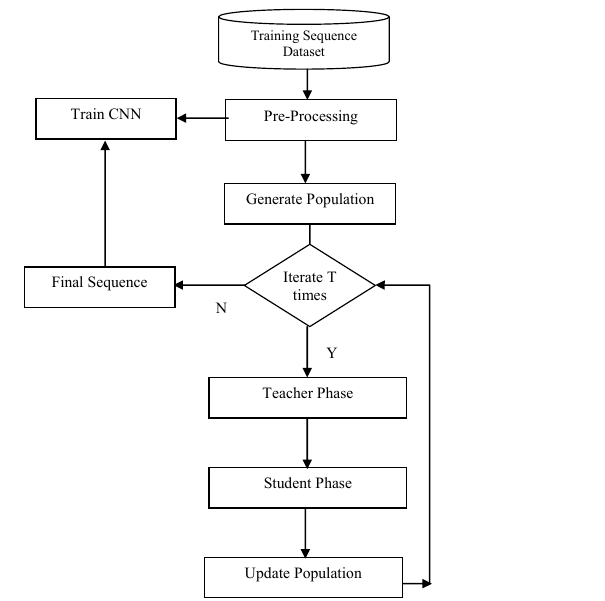}
    	\caption{CNNLB-TLBO Flowchart} 
\end{figure}
\hspace*{1em}At the same time, some talent researchers design a neural network-based scheduling optimization plan when studying the optimization strategy for the hospital emergency department \cite{rosemarin2019emergency}. The program's primary purpose is to automatically allocate and dispatch medical staff to shorten the gap in patient care and improve the emergency room's performance indicators without changing the existing emergency room arrangements. In addition, referring to the research on neural network sorting methods by Rigutini et al. \cite{rigutini2011sortnet}, these researchers finally propose a new type of neural network shown in Figure 4 \cite{rosemarin2019emergency} to realize the forecast. \\
\begin{figure}[H] 
    	\centering
    	\includegraphics[width=5cm,height=3cm]{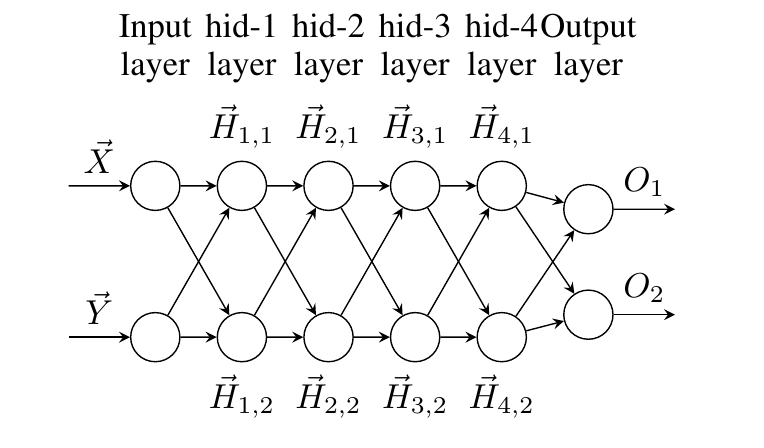}
    	\caption{Neural Network for Hospital Scheduling} 
\end{figure}
\section{Mathematic Model Construction}
The manpower scheduling problem is an NP-HARD problem, which means that we cannot find the most effective solution in polynomial time. Under normal circumstances, the scheduling problem's solution is to build a mathematical model and use heuristic algorithms to solve the objective function of the mathematical model, thereby obtaining the optimal or suboptimal solution to the scheduling problem. The steps proposed to solve the objective function are summarized in the data acquisition part of this paper.\\
\hspace*{1em}We firstly build the model to solve basic manpower scheduling problems. If we set the input data as $x$ and express the $K$ constraints as \(\varphi_1,\varphi_2,…,\varphi_K\), then the objective function of the scheduling problem demand is expressed as \(f(x|_{\varphi_1,\varphi_2,…,\varphi_K} )\). Normally, the input data $x$ is a matrix that contains the number of days required for the scheduling task and the type of work that needs to be scheduled, which are randomly initialized. Our goal is to solve the $f(x|_{\varphi_1,\varphi_2,…,\varphi_K})$ and obtain the corresponding $x \in R^n$ under the premise of satisfying the constraints. The result $x$ obtained after the solution can exist as a set of solution spaces. After solving the personnel arrangement, we still need to design an algorithm to realize the specific output personnel arrangement plan. The main task in this step is to arrange the precise positions for every worker every day. Therefore, the domain parts of the model in this paper mainly include establishing constraint conditions, the choice of the objective function, the calculation, and the expression of the solution space. \\
\begin{figure}[htbp] 
    	\centering
    	\includegraphics[width=6cm,height=6cm]{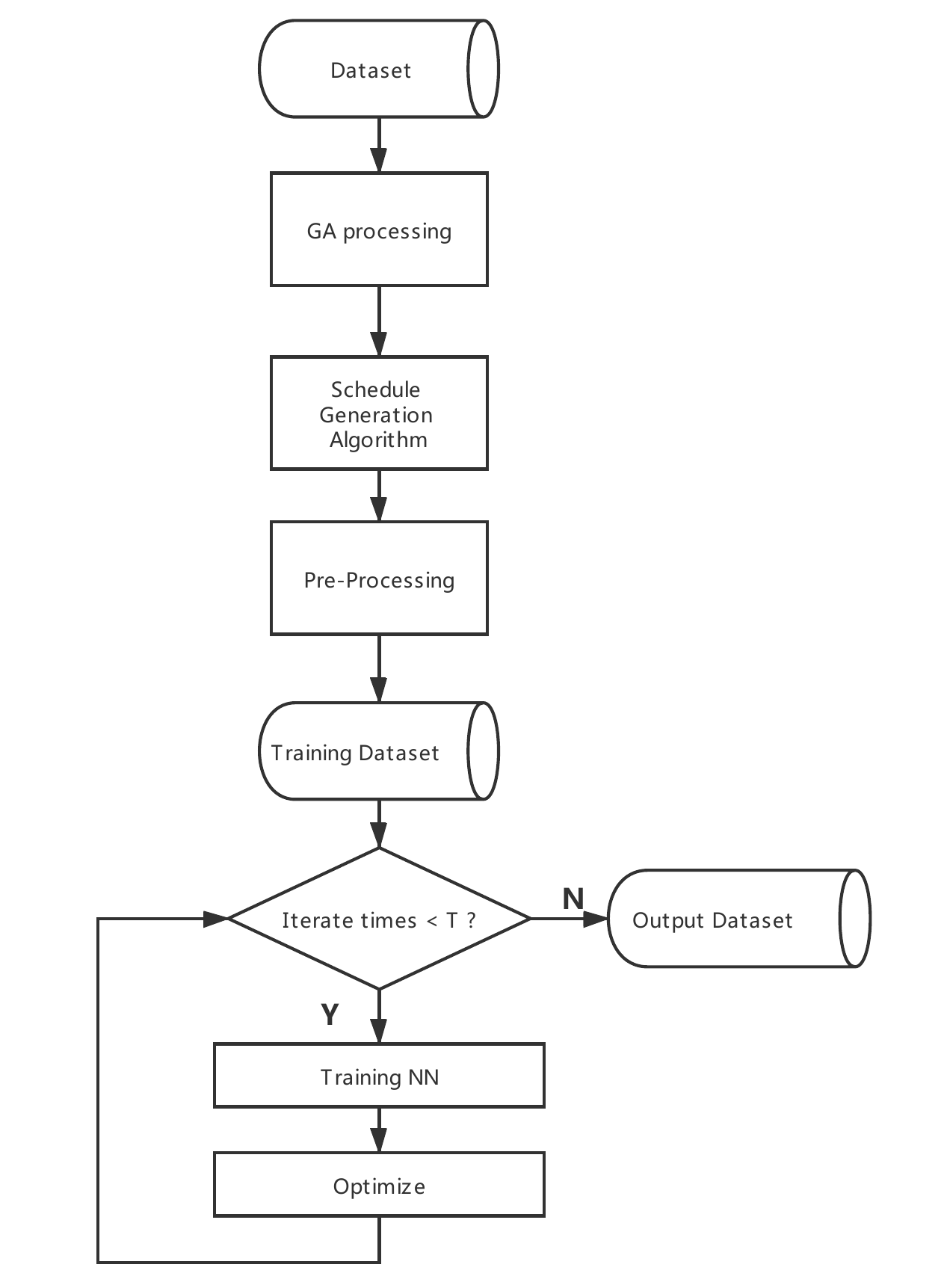}
    	\caption{Optimal model with DNN} 
\end{figure}
\hspace*{1em}Inspired by the research from Rosemarin, Rosen-feld, and Kraus \cite{rosemarin2019emergency}, we are confident in the idea that neural network forecast for scheduling problems can be migrated and expanded. Using the well-designed genetic algorithm, we can quickly solve the scheduling plan for one week. However, when the scheduling cycle becomes longer, the actual working time complexity of the algorithm that we use to run the loop will become extremely long, which also leads to a decrease in the efficiency of the algorithm. Therefore, we need to change $\dim(x)$ so that $x \in R^n$ becomes $y \in R^m$, and $y$ can be used to train the neural networks. We will use the scheduling results within a week as the training set and use the feedforward neural network to forecast scheduling arrangements in the future. The basic process of the optimized scheduling scheme we design is using the genetic algorithm and the scheduling generation algorithm to obtain the scheduling results within a period, and then create a neural network to train the input-output data set and compare it with the test set. Finally, we design experiments to check the accuracy of the chosen model. The specific workflow of our model is shown in Figure 5.\\
\subsection{Constraints and Objective Functions}
\hspace*{1em}Different companies have different business logic, business rules, working conditions, et cetera. Therefore, the more options for the constraints of the scheduling problems we have, the better our forecast model will perform. Taking into account the requirements of most companies on the market \cite{tambe2019artificial} and the labor law \cite{bai2020employment}, this paper summarizes the commonly used scheduling constraints as follows:\\
\begin{itemize}
\item $\varphi_1$: Every employee has a fixed job.
\item$\varphi_2$: It must be ensured that every post is on duty every day.
\item$\varphi_3$: There are upper and lower limits on the working hours within one cycle.
\item$\varphi_4$:There are upper and lower limits on the total salary of employees. 
\item$\varphi_5$: There are upper and lower limits on the total number of employees.
\item$\varphi_6$: There are lower limits on holiday periods for employees. 
\item$\varphi_7$: Priority should be assigned for urgent tasks. 
\item$\varphi_8$: There are upper and lower limits on the number of employees. 
\item$\varphi_9$: Schedule in a specific order.
\item$\varphi_{10}$: Provide arrangements to meet shift needs. 
\item$\varphi_{11}$: Cooperation of different jobs for a specific task. 
\end{itemize}
\hspace*{1em}Different companies often require different logical combinations. The logical combination here refers to the use of logical connectives $\wedge, \vee, \lnot$ to connect constraints in series to generate what we need.\\
\hspace*{1em}The objective function selection of common scheduling optimization problems can generally be summarized into the following three types:\\
\begin{itemize}
\item Minimize the number of employees $\min \sum_{i=1}^{n} x_i$
\item Minimize total working time for employees $\min T_{total}$
\item Minimize the cost of total salary $\min C_{cost}$
\end{itemize}
\subsection{The Construction of Problems}
According to the constraints and the objective function, we can construct the scheduling optimization problem within the framework proposed in this paper. For the most basic set of scheduling problems, that is, “To meet the requirements of each position, there must be no interspersed on-duty, and there must be someone on duty at all times during working hours. In addition, the working hours, vacation time, and wages of employees must be within a certain range. The target is to arrange a scheduling plan with the smallest total time for employees.”, can be expressed as:
\begin{equation}
\left\{\begin{array}{ll}x_{best} = \arg \min \limits_{x} T_{\text {total}} \\ \text {subject to } \varphi_{1} \wedge \varphi_{2} \wedge \varphi_{3} \wedge \varphi_{4} \wedge \varphi_{5} \wedge \varphi_{6} = True\end{array}\right.
\end{equation}
\hspace*{1em}By using heuristic algorithms such as genetic algorithm \cite{dauphin2014identifying}, simulated annealing algorithm \cite{wirawan2015lp}, et cetera. to solve the optimization problem, we can obtain the minimum total time and the corresponding total demand of employees Quantity.
\subsection{Generation algorithm for manpower scheduling problems}
In order to create the training set, we have designed the following algorithm to achieve the requirements of generating the corresponding scheduling results under the suitable constraints affected by constraints and other factors.\\
\hspace*{1em}The WorkerList collection is used to store the results of the arrangement for employees; the WorkableSet collection is used to measure the attendance of employees; Daylimit is the number of working days; the WorktimeSet collection is used to count the working hours of employees; the RequiredNumber is used to indicate the number of needs of employees in the corresponding position, which we can obtain from the results of solving the optimization problem. \\
\hspace*{1em}For more complex scheduling problems, we need to create a function $Suitable()$ that determines whether employees are suitable for joining the scheduling plan. When the employees we obtain do not meet our scheduling constraints, we also create a function $ChangeOrder()$ to reconstruct the selection of employees. In summary, our complete scheduling algorithm is:
\begin{breakablealgorithm}
\small
\caption{Improved Scheduling Algorithm}
\begin{algorithmic}
\STATE{Initialize $WorkerList$, $WorkableSet$, $Daylimit$, $WorktimeSet$, $RequiredNumber$, $Proficiency$, for employees}
\STATE{Initialize $Constraints$}
\STATE{Initialize $Counter1$ as 0}
\WHILE{$Counter1<Daylimit$}
\STATE{Initialize $Counter2$ as 0}
\STATE{Initialize $Container$}
\WHILE{$Counter2<RequireNumber$}
\STATE{assign $Man$ using $random(Memberlist)$}
\IF{$Suitable(Man,Constraints)$ \textbf{is} $True$}
\STATE{Add $Man$ into $Container$}
\STATE{Increase $WorkableSet[man]$ and $Counter2$}
\ELSE
\STATE{assign $NewMan$ using \\$ChangeOrder(Man,WorkerList,WorkableSet)$}
\IF{$Proficiency[Man]$ $\ge$ $Proficiency[NewMan]$}
\STATE{Add $Man$ into $Container$}
\STATE{Increase $WorkableSet[Man]$ and $Counter2$}
\ELSE
\STATE{Add $NewMan$ into $Container$}
\STATE{Increase $WorkableSet[NewMan]$ and $Counter2$}
\ENDIF
\ENDIF
\ENDWHILE
\STATE{Add $Container$ into $WorktimeSet$}
\STATE{Increase $Counter1$}
\ENDWHILE
\end{algorithmic}
\end{breakablealgorithm}
\subsection{Optimization for Neural Network}
The normalization method selected in this paper is MinMax standardization \cite{munkhdalai2019mixture}. Moreover, we choose to set up five different neural networks and analyze the training results from them. The five different networks are Forward Deep Neural Network (FDNN), Recurrent Neural Network (RNN), Radial Basis Function Neural Network (RBFNN), Long Short-Term Memory Network (LSTM), Gate Recurrent Unit (GRU), and Convolutional Neural Network (CNN). Moreover, we choose Pytorch \cite{pytorch17} as the frame for building neural network architecture. Pytorch is an open-source machine learning library based on tensors and developed by Facebook's artificial intelligence team, and its high quality has been proved by researchers \cite{paszke2019pytorch}.
\begin{table}[H]
\caption{Parameters for Different Networks}
\begin{center}
\small
\setlength{\tabcolsep}{0.1mm}{
\begin{tabular}{|c|c|c|c|}
		\hline
		 Name &Number of input units&Number of layers&Activation function\\ \hline
		FDNN&32&4&Sigmoid\\ \hline
		CNN&32 or 4&10&Sigmoid\\ \hline
		RBFNN&32&3&Gaussian\\ \hline 
		RNN&4&10&Tanh\\ \hline
		LSTM&4&10&Tanh\\ \hline
		GRU&4&10&Tanh\\ \hline
\end{tabular}
}
\end{center}
\end{table}
\begin{table}[H]
\caption{Activation \& Cost functions}
\begin{center}
\footnotesize
\setlength{\tabcolsep}{0.00000001mm}
{
	\begin{tabular}{|c|c|}
		\hline
		\multicolumn{2}{|c|}{Activation functions}\\
		\hline
		Sigmoid & $f(x)=\frac{1}{1+e^{-x}}$ \\ 
		\hline
		Gaussian & $f(x)=e^{\left(-\frac{\left\|x-\mu_{i}\right\|^{2}}{2 \sigma^{2}}\right)}$ \\ 
		\hline
		Tanh & $f(x)=\frac{e^{x}-e^{-x}}{e^{x}+e^{-x}}$ \\ 
		\hline
		\multicolumn{2}{|c|}{Cost functions}\\
		\hline
		MSELoss & \(L=\left\|y_{n}-\hat{y}_{n}\right\|^{2}\) \\ \hline
		L1Loss & \(L=\left|y_{n}-\hat{y}_{n}\right|\) \\ \hline
		SmoothL1Loss & \(L=\left\{\begin{array}{c}\frac{1}{2}\left(y_{n}-\hat{y}_{n}\right)^{2} \text { for }\left|y_{n}-\hat{y}_{n}\right| \leq \delta \\ \delta\left|y_{n}-\hat{y}_{n}\right|-\frac{1}{2} \delta^{2} \text { else }\end{array}\right.\) \\ \hline
		BCEWithLogitsLoss & \(L=w\left[y_{n} \log \left(\sigma\left(\hat{y}_{n}\right)\right)+\left(1-y_{n}\right) \log \left(1-\sigma\left(\hat{y}_{n}\right)\right)\right]\) \\ \hline
	\end{tabular}
}
\end{center}
\end{table}
\hspace*{1em}The following is a brief explanation of the choices of different neural networks. The feedforward neural network FDNN is a relatively easy neural network to build, and its training methods are more diverse and mature. The confidence of the output results is also high. The reason for choosing the recurrent neural network RNN is using the scheduling problem's timing series. Proven by the work from He et al. \cite{he2019gold}, using the LSTM-CNN neural network to predict gold price trends, we believe that the time series forecasting model can be used to solve the scheduling problem since the results of the scheduling plan are well-regulated. RNN, LSTM, and GRU can be used for related forecasting work. RBFNN has been used by some researchers \cite{TRIPATHY2015101} to solve similar scheduling problems, so we will test the effectiveness of the network again. Convolutional neural network CNN is usually used in text or image processing, but the convolution can also collect information from previous data and give a forecast. Finally, we will also compare the effects of the five neural networks in the experiment to determine the outstanding one.\\
\hspace*{1em}For the accuracy of our prediction result, referring to the target of manpower scheduling problem, we define the variable $V_{cc}$ as: 
\begin{equation}
V_{cc}=\frac{M_{t}}{Days}
\end{equation}
\hspace*{1em} $M_{t}$ means the time for correct matching between the prediction result and test data, and $Days$ means the length of time series.\\
\hspace*{1em}For the selection of neural network-related parameters, we need to compare the convergence speed. The parameters we choose can make the neural networks earn the relatively highest convergence speed. Specifically, CNN contains two convolution layers and eight linear layers. Since the scheduling of the solution to the scheduling dimension, we first consider encoding the independent input variables for the problems involving time to enlarge the space of information. Since the difference between the input of days when expressed in decimal is not significant enough, and the amount of information is not sufficient, we choose to convert all the input data of FDNN, RBFNN and CNN into 32-bit binary numbers. This transformation can increase the amount of input information, and the forecast results will be more accurate. For the input data of RNN, LSTM, and GRU, we will need to normalize input and adjust time steps because we adopt the method of time series forecast. Table I and Table II contain more information about this.\\
\hspace*{1em}In the experiment part, this paper adopts the Adamax algorithm \cite{kingma2014adam} as the optimization algorithm. This algorithm can update the learning rate adaptively and it is one of the best optimization algorithms. Cost function chosen by this paper is the Mean Square Error function (MSE).
\section{Experiment}
In this part, we solve two different scheduling problem using our model. The first problem is to arrange jobs scheduling in one big market, and we take cashier position to justify six different structures. The supremum of total employees is 60. After 2000 iterations, we can observe the error curves and accuracy curves as Figure 6\&7 shows.
\begin{figure}[H] 
    	\centering
    	\includegraphics[width=7cm,height=5cm]{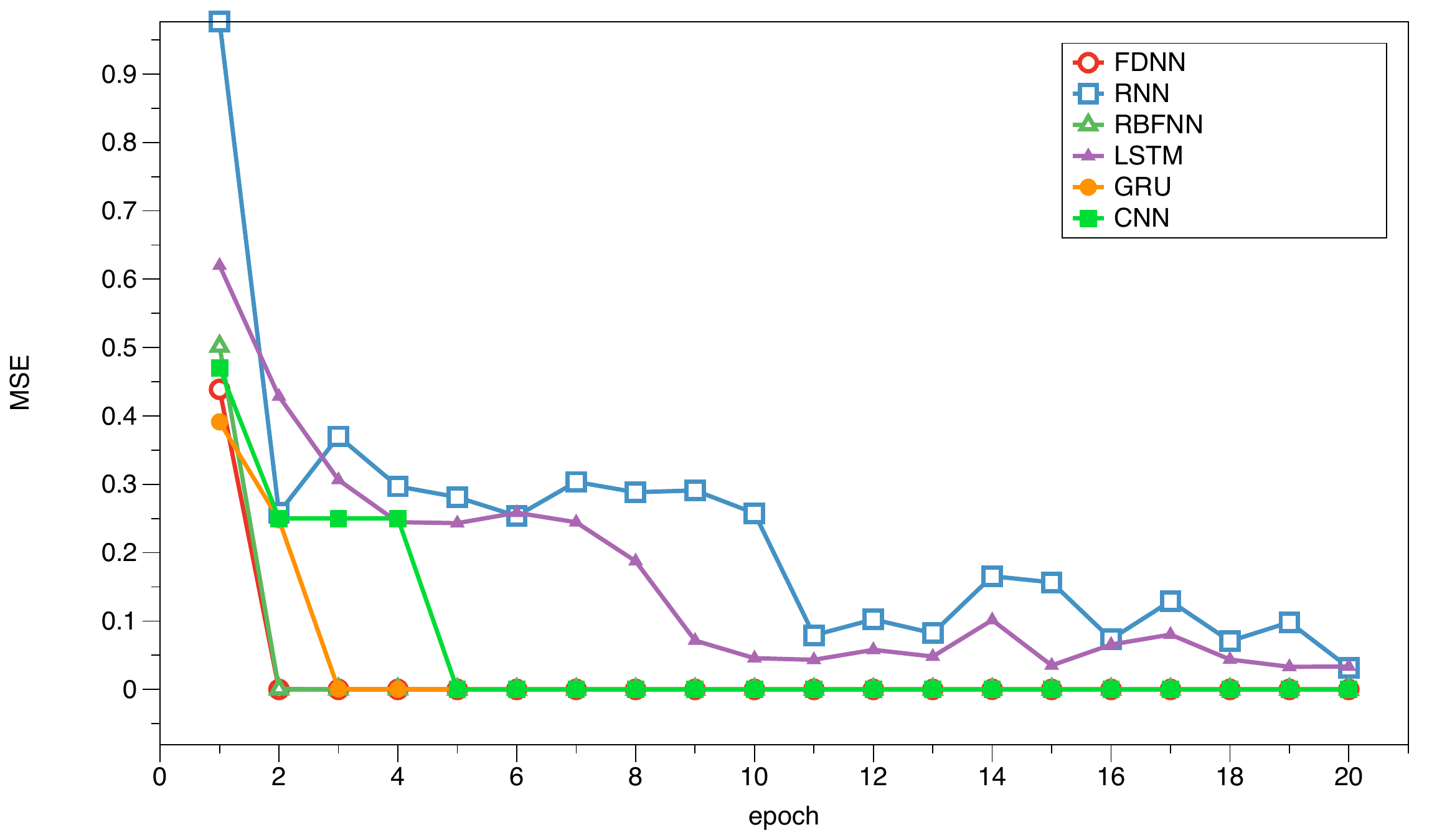}
    	\caption{MSE of Different Networks} 
\end{figure}
\begin{figure}[H] 
    	\centering
    	\includegraphics[width=7cm,height=5cm]{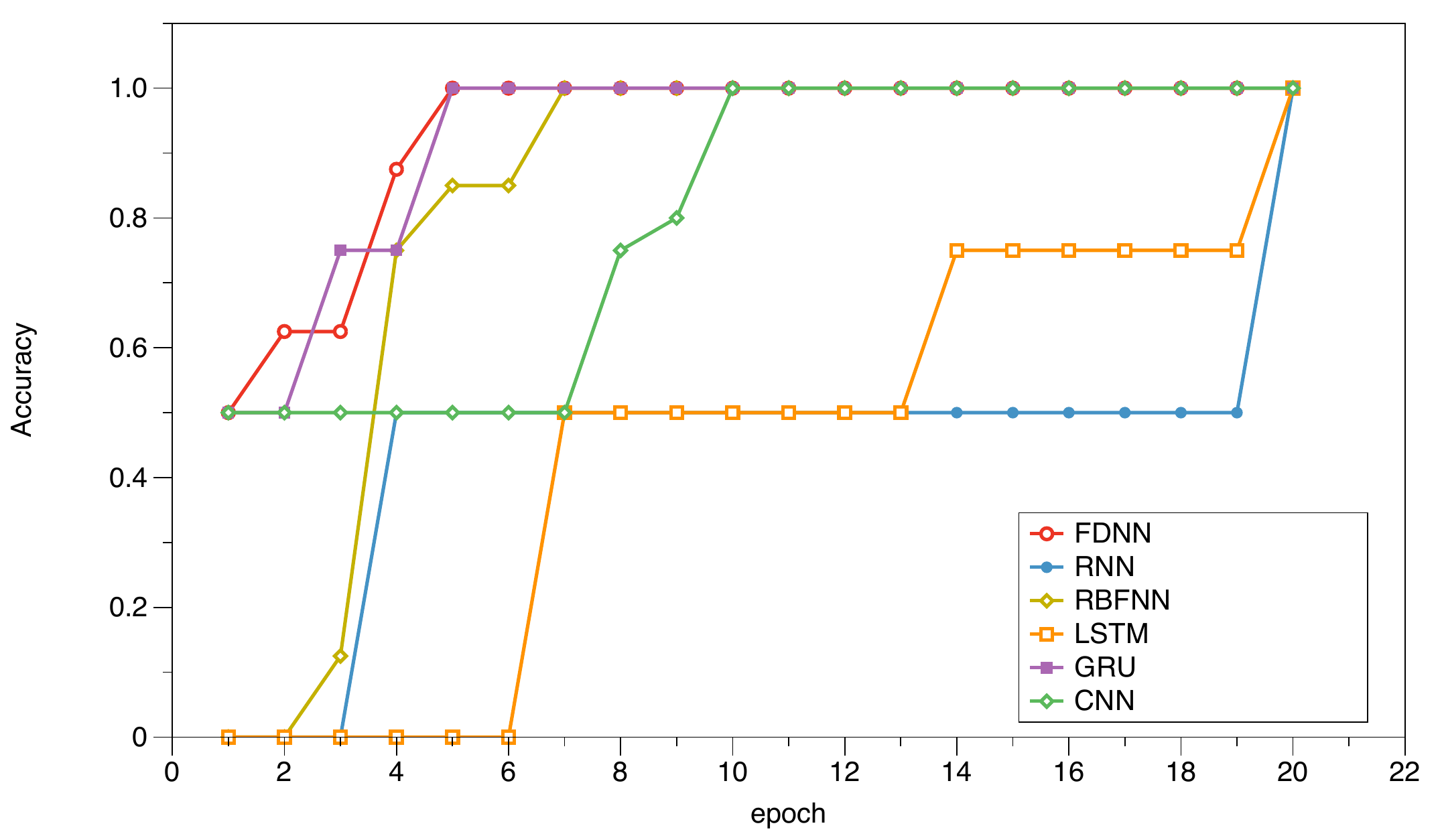}
    	\caption{Accuracy} 
\end{figure}
From the test results, we can see that our FDNN, RBFNN, GRU, LSTM, CNN are the models with better convergence and the more precise forecast results. In the next step we can use the neural network training method to select the best neural network among FDNN, RBFNN, CNN, LSTM, and GRU for further tasks. For the position clerk, in the scheduling task, the performance of five kinds of neural network training is shown in Figure 8.
\begin{figure}[H] 
    	\centering
    	\includegraphics[width=8cm,height=5cm]{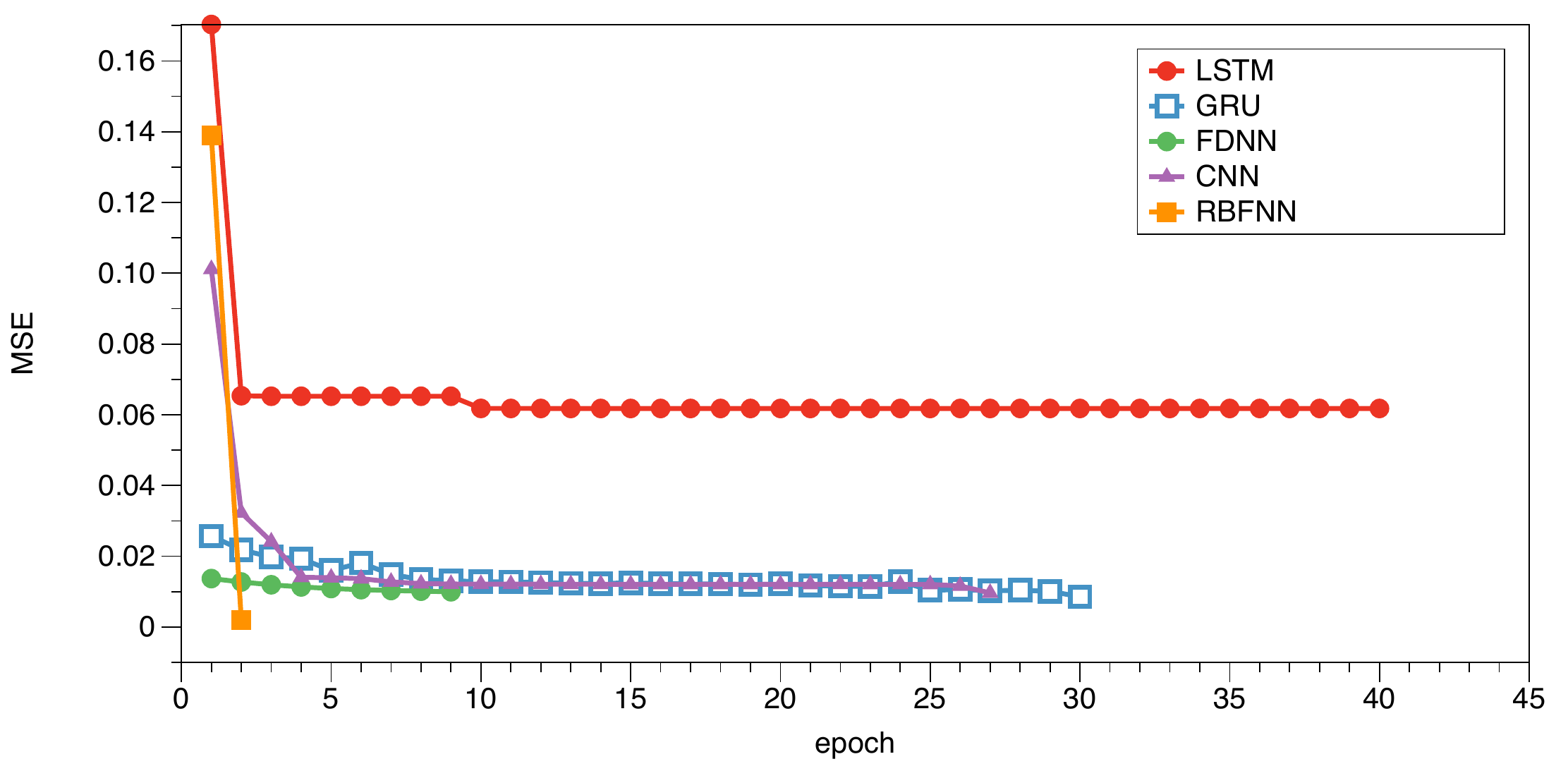}
    	\caption{MSE of Different Networks} 
\end{figure}
\begin{figure}[H] 
    	\centering
    	\includegraphics[width=8cm,height=5cm]{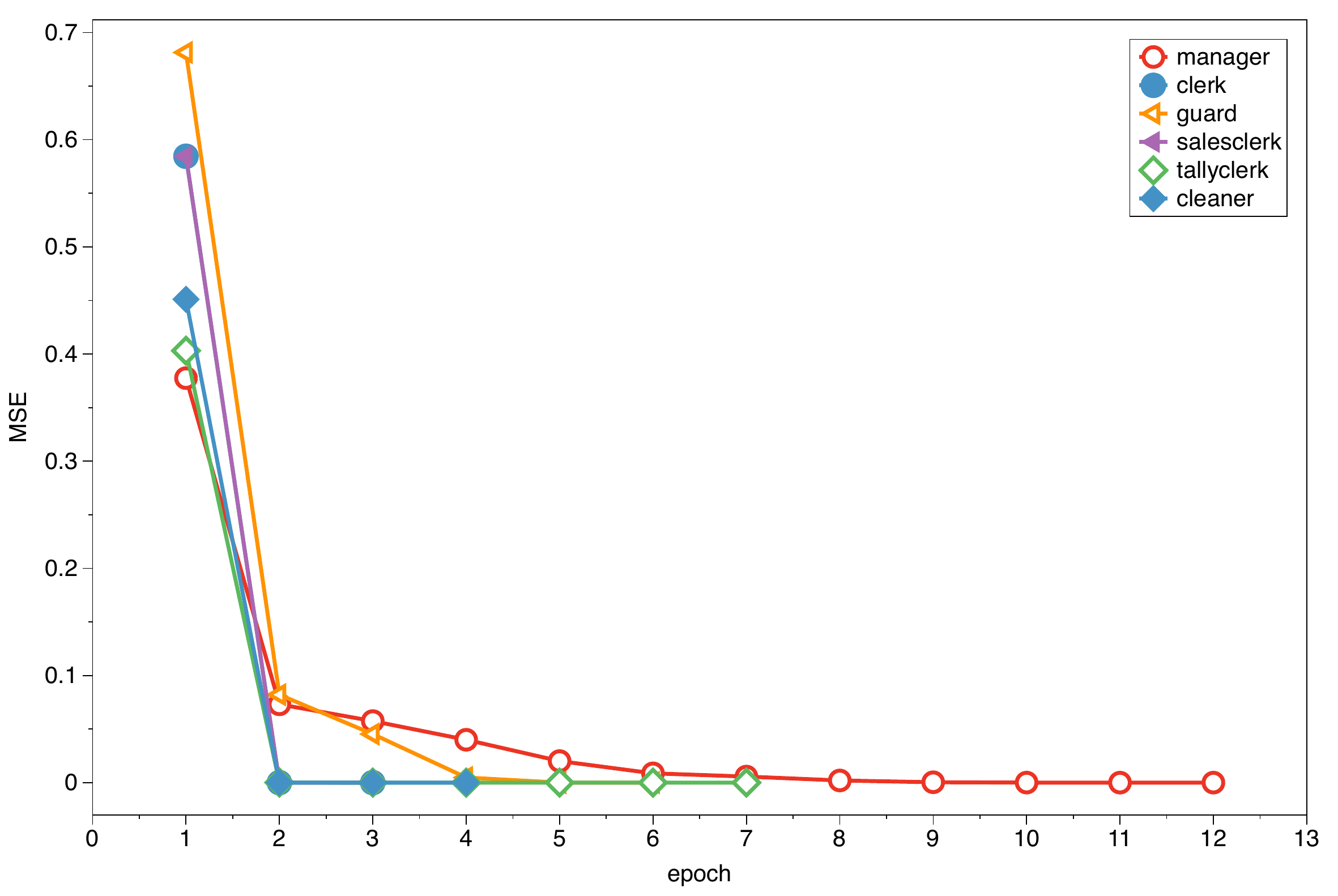}
    	\caption{MSE for Different Employees} 
\end{figure}
From the results, we can see that the neural network with fastest convergence rate is RBFNN, However, there is a significant risk of over-fitting in such network, so we will drop it. Moreover, FDNN is more suitable network with relatively fast convergence speed and better accuracy. LSTM has slow convergence speed and sometimes it will be trapped in the process and stop optimizing. This situation happens many times during our experiment. Therefore, we use our FDNN to complete the forecast of complex scheduling schemes within one month for different jobs. After setting the termination error at \textbf{1e-7}, the training result of the FDNN is in Figure 9.
The experimental results show that the FDNN optimized by Adam algorithm can fit and predict the scheduling of different positions well.\\
\hspace*{1em}The other problem is the issue of bus scheduling, presented and discussed by Lucas Kletzander\cite{Kletzander_Musliu_2020}. Based on the constraints provided by this paper, we use our model to generate a scheduling table for 8 buses. The chosen network is FDNN, the result of errors is shown in Figure 10, the training result and arrangement result is shown by the Figure 11. In this figure, the value of points in arrangement axis can only be 0 (No attendance) or 1 (Attendance). In the arrangement result, our model generates the scheduling table for different buses, which can be utilized in real world arrangement. This experiment demonstrates the feasibility of our use of a neural network-based mathematical model to solve generalized scheduling problems and also illustrates the potential and promise of our research.
\begin{figure}[htbp] 
    	\centering
    	\includegraphics[width=8cm,height=6cm]{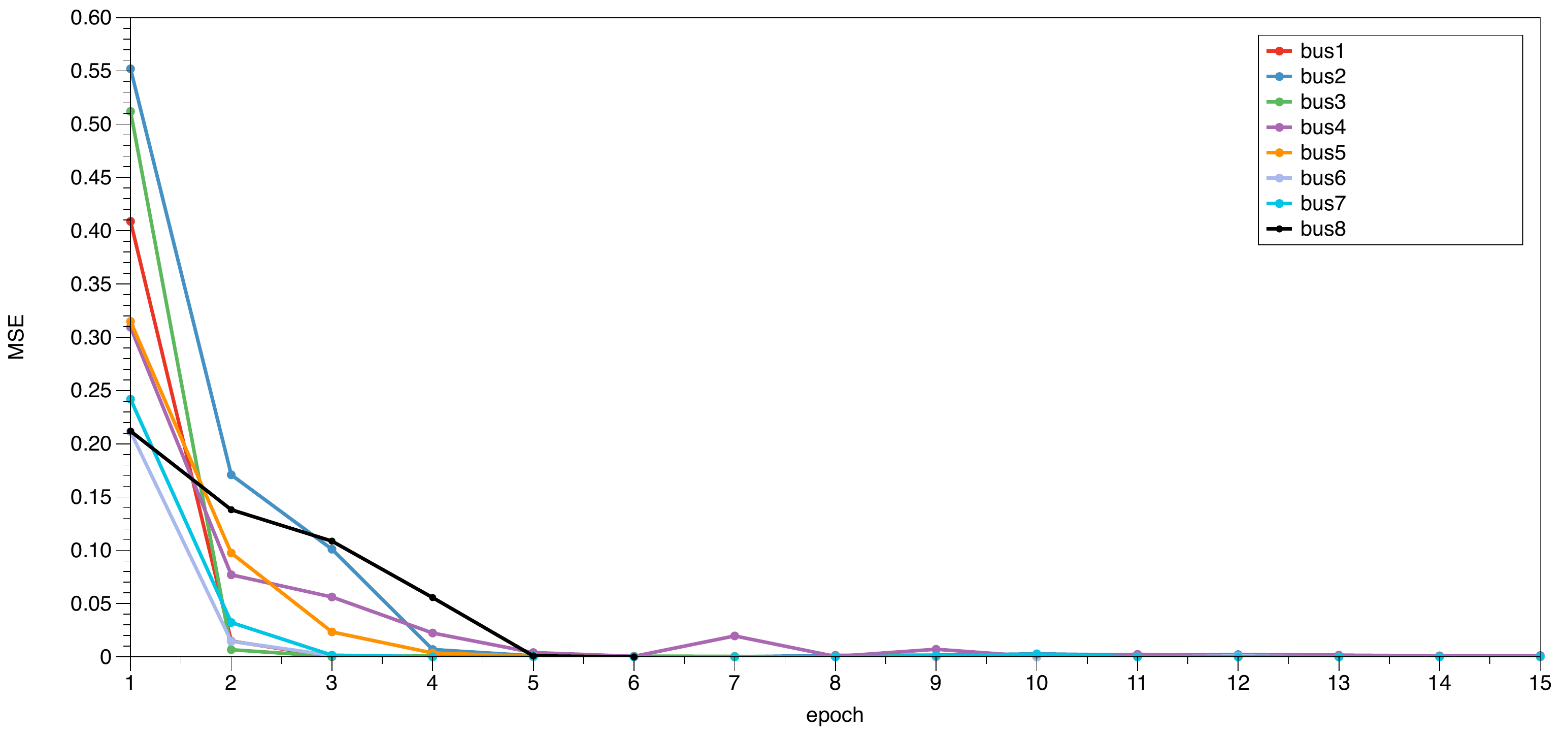}
    	\caption{MSE of Bus Arrangement} 
\end{figure}
\begin{figure}[htbp] 
    	\centering
    	\includegraphics[width=8cm,height=6cm]{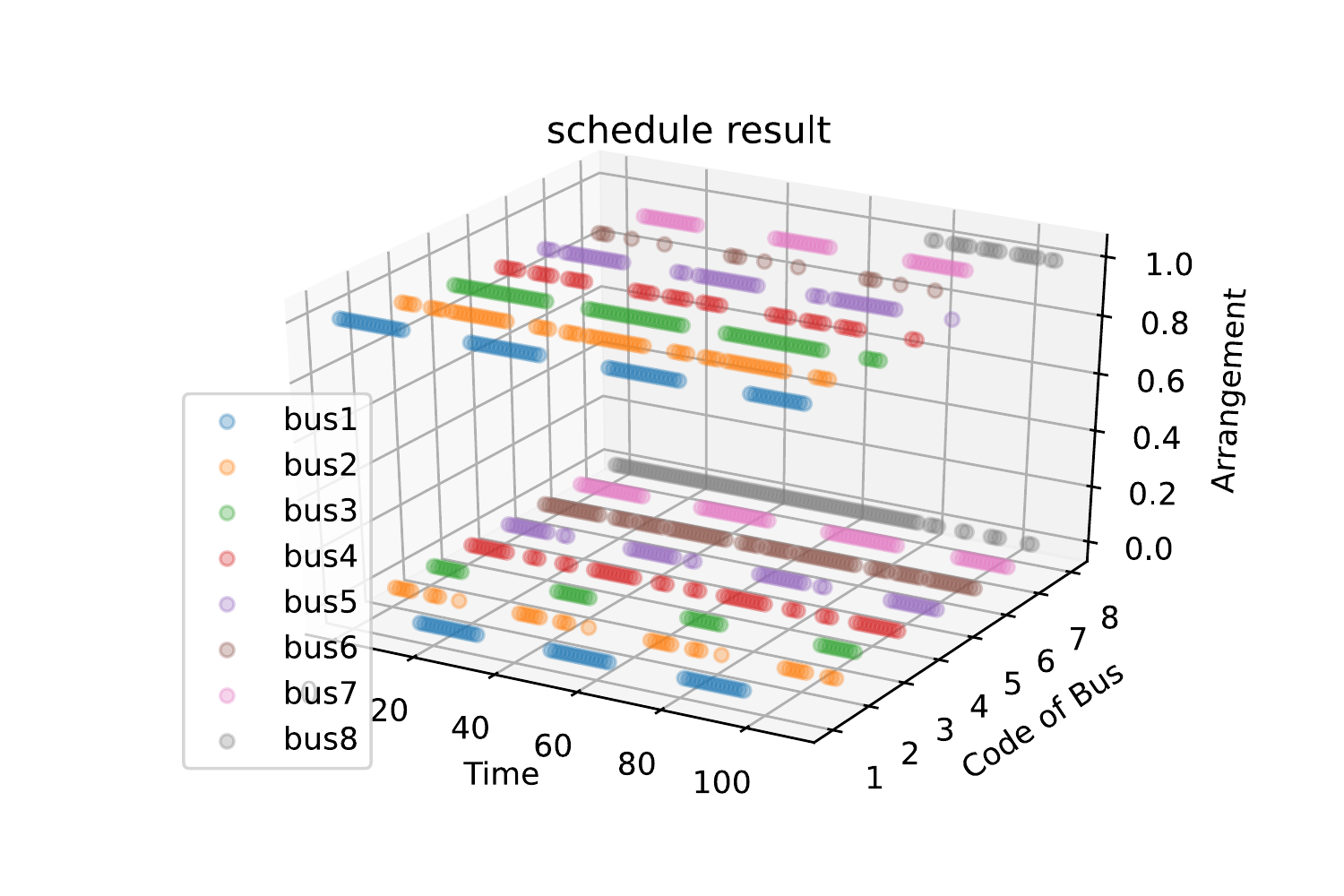}
    	\caption{Bus Arrangement} 
\end{figure}
\section{Discussion}
This paper discusses the errors from the following three points: 1. The influence of neural network structure. 2. The effect of different optimization strategies.\\
\hspace*{1em}Different neural networks adopt different learning methods and forecast methods. According to the universal approximation theorem \cite{hornik1989multilayer}, for FDNN, the structure of the neural network does not have a great impact on the results. Because the universal approximation theorem shows that if a feedforward neural network has a linear output layer and at least one hidden layer with any kind of “squeezing”activation function (such as logistic sigmoid activation function), it only needs to give the network a sufficient number The hidden unit of, it can approximate any Borel measurable function from one finite dimensional space to another finite dimensional space with arbitrary precision. Borel function is a continuous function in compact set, and any continuous function defined on the bounded set of $R^n$ is Borel measurable. The theorem shows that the neural network can approximate any continuous functions in $R^n$  by arbitrary accuracy, and this work has been further pushed by other researchers \cite{lin2017does}. Moreover, we use the activation functions such as $Sigmoid$ and $ReLU$ which have a "squeeze" feature to build the neural networks, so the forecast result is within an acceptable error range. For RNN, LSTM, and GRU, the time series forecast increases the probability of searching for the global optimal solution. The output results of these three types of neural networks still meet the requirements. However, the neural networks that use time series forecast has a slower convergence speed, and if time series forecast is applied to CNN, the accuracy or the optimizing of loss will become worse. Therefore, time series forecasting requires a more stringent neural network structure than using time as standard input in training networks.\\
\hspace*{1em}In our opinion, the optimization strategy is most likely to affect the neural network's forecast results (Figure 12). The choice of optimization strategy includes optimization methods and cost functions. When the optimization method selected in this paper encounters the saddle point, although the value of loss will be stable, the result is not optimal. Regarding the influence of the saddle point on the optimization process, Dauphin et al. \cite{dauphin2014identifying} have proved that the main factor that causes the neural network to fall into the larger error trap is the saddle point, not the optimal local solution. Moreover, when caught in such a dilemma, the Adam algorithm has no stable approach to solve it, so we first consider choosing other types of optimization solutions. For model optimization methods, we can choose AdamW \cite{loshchilov2018fixing}, Adamax \cite{kingma2014adam}, RMSprop \cite{tieleman2012lecture}, and other similar methods. In addition, We can also choose different cost functions to adjust the standard of measuring loss to better determine the accuracy of the model. 
\begin{figure}[H] 
    	\centering
    	\includegraphics[width=7cm,height=5cm]{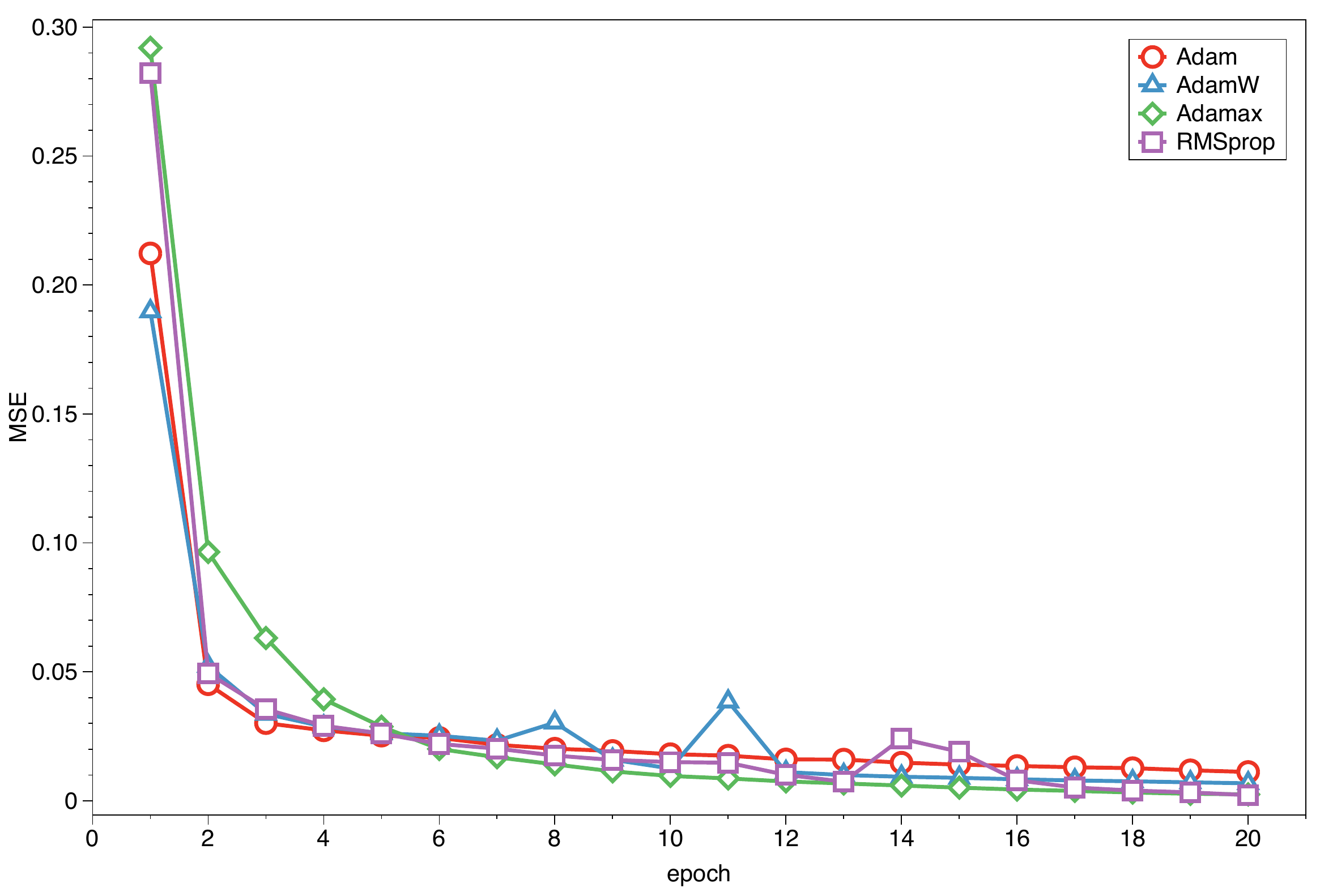}
    	\caption{Different Optimization Strategies} 
\end{figure}
\begin{figure}[H] 
    	\centering
    	\includegraphics[width=7cm,height=5cm]{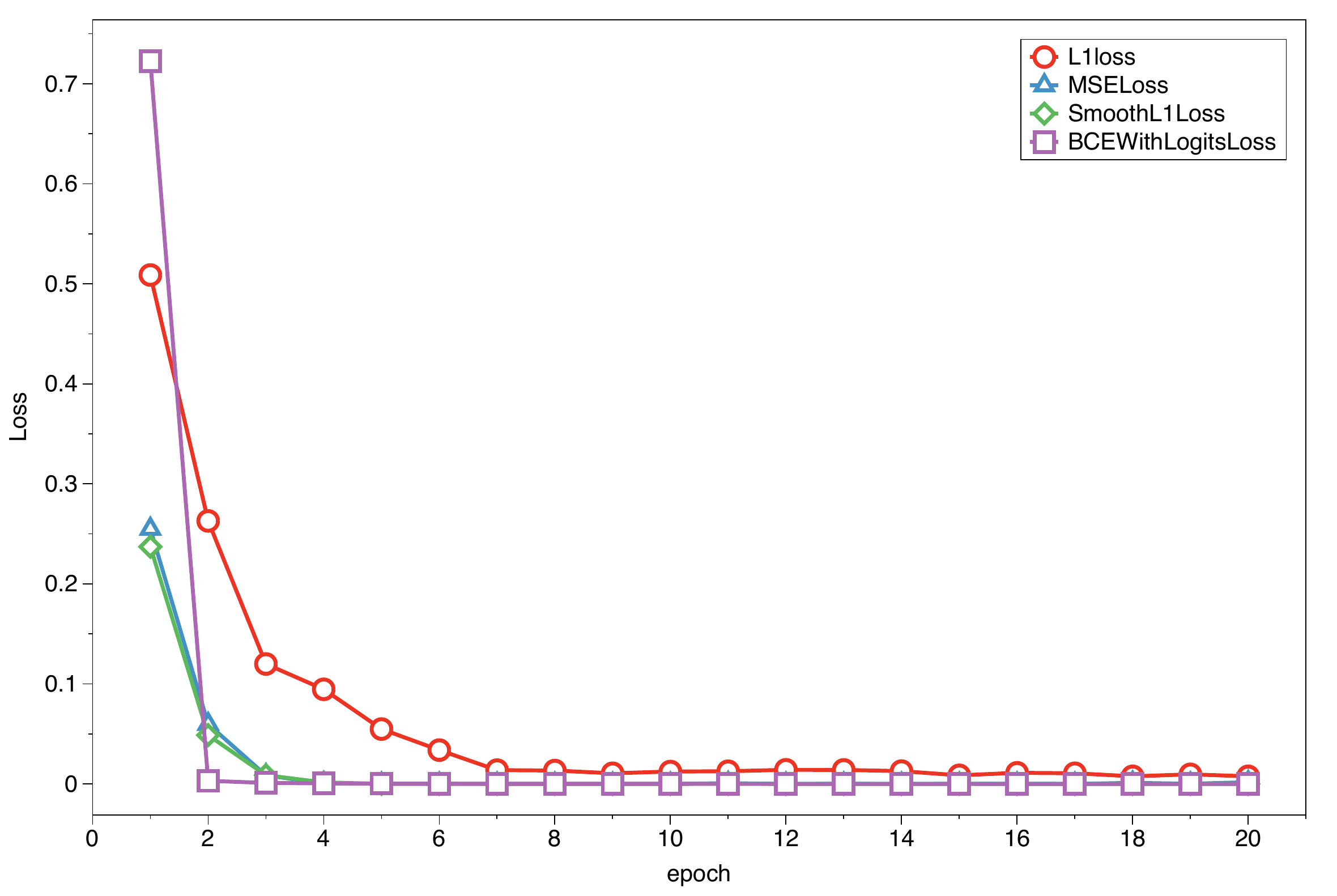}
    	\caption{Different Cost Functions} 
\end{figure}
From the results of different strategies, we can see that Adamax and RMSprop are better than the Adam algorithm in actual use. Our explanation for this is that Adamax and RMSprop appropriately set up a more flexible boundary (upper limit) of the learning rate. Therefore, when a saddle point or optimal local situation is encountered, the probability of using these two algorithms to jump out of the trap will be higher, thereby ensuring the model continues to be optimized.\\
\hspace*{1em}From the results of different cost functions (Figure 13), we can also conclude that we select a suitable cost function for our neural networks under certain conditions. However, adjusting the type of cost function on the final result is not very significant. Therefore, we can preliminarily determine that the cost function's choice is not the cause of significant errors in some neural network results.\\
\hspace*{1em}It can be concluded that the arrangement of the data set, and the saddle points in the model training are most likely to affect the neural network's forecast results.
\section{Conclusions}
By exploring the scheduling problem method, this paper uses two specific scheduling problem-solving processes to illustrate a model that combines a neural network to give forecasts for the scheduling plan. The scheduling problem is a combinatorial optimization problem composed of objective functions and constraints. For the basic scheduling requirements, this paper combines the genetic algorithm with combinatorial optimization so that we will obtain the heuristic algorithm's output results. These output results can realize the generation of scheduling tasks in the next few days and be used for neural network training. In the next step, we compare the effect of five different neural networks so that we can determine four neural networks with high fitness and perform well in the dataset. For more complex scheduling requirements, the modified heuristic algorithm and the scheduling generation algorithm's structure are required. The new results can be used for further training to determine the best network-Deep Feedforward Neural Network. In total, the heuristic algorithm has played a leading role; the scheduling task generation algorithm has played a mainstay role; the neural network model is the core of the forecast structure and has played the role of outputting the correct results. This paper proves that using theses three algorithms (or methods) together as a model will make it possible to solve the scheduling problem more efficiently.\\
\hspace*{1em}In addition, this paper also conducts experiments and discussions on the diversity of machine learning methods/neural network structures. For the selection of different neural networks, this paper proves the convergence and accuracy of the feedforward deep neural network (FDNN) results; for the selection of different cost functions, this paper proves the rationality of choosing the MSE cost function. As for the reasons affecting errors in the model, this paper also conducts in-depth exploration, looking for the source of the errors from the aspects of network structure, and optimization strategy selection. After a detailed discussion, we finally conclude that the form of data input and the choice of optimization plan are two critical factors that affect the rationality of the algorithm's output from the scheduling problem. Therefore, researchers should invest time and resources in data preprocessing schemes and the generation of advanced optimization algorithms.
\section{Furture Work}
In the future, we plan to develop a more advanced neural network structure, which is based on time series forecasting to shorten the task workflow and the time cost for the generation. We are also interested in comparing with more models to collect more data.
\section{Acknowledgments}
We are grateful to anonymous reviewers for their helpful comments. This work is supported in part by NSFC Grant No. 61772315 and 61861136012, and National Key R\&D Program of China (No. 2018AAA0101100).
\bibliographystyle{IEEEtran}
\bibliography{references}

\end{document}